# Reasoning with Uncertain Knowledge

A. Julian Craddock & Roger A. Browse

Department of Computing and Information Science
Department of Psychology
Queen's University at Kingston
Ontario, Canada

Heuristics derived from human experimentation are integrated with a knowledge network to produce a model for reasoning with uncertain information. The believability of knowledge is determined by collecting reasons for believing and not believing it. These reasons, or endorsements, are subsequently ordered by their belief, the reliability of their belief, and their importance relative to one another.

**INTRODUCTION**: The development of mechanisms for the representation of knowledge has always been a central concern of artificial intelligence. The fundamental criteria for representational schemes have been adapted from criteria mathematics has set up for logical formalisms: (1) that a translation into the representation from natural language statements must be possible, and (2) that deduction and inference of a sort that yields results similar to human conclusions must be possible over the representation. These criteria are not well met by knowledge representation schemes which are based on traditional mathematical logic. In particular human expression is characterized by the use of measures of uncertainty, and human reasoning often appears to not follow the dictates of logic and probability (Lindley, 1971).

There are several solutions for the problems of uncertain information. The first is what Cohen (1983) called the "engineering solution". This solution is used by many models in artificial intelligence (McDermott and Doyle 1980; McDermott 1980; McCarthy 1979; Reiter 1980). The solution does not deal with uncertainty as a useful source of information and constraints; instead it reduces the problem domain in such a manner as to eliminate uncertainty. Unfortunately eliminating uncertainty results in a reformulated problem that is, at the best, only vaguely related to the original. The second solution is to make quantitative assumptions about uncertainty using probabilities and possibilities (Zadeh 1983, 1984; Lee 1969; Edwards 1982; Shortcliffe 1975). The quantitative assumptions often prove to be overly restrictive and lacking in expressive power. The third solution is to use a utility based solution (Nosteller and Nogee 1951; Schoemaker 1980). This solution makes the unpleasant assumption that we can determine subjective utilities for events and manipulate these utilities in a formal manner. Also, there are no clear indications that humans attempt to maximize their expected utilities while reasoning.

A more promising approach to the role of uncertainty in human reasoning is presented by Kahneman and Tversky (1982a b). Their model indicates that humans employ a set of basic heuristics which aid in making decisions in conditions of uncertainty. These heuristics enable humans to constrain problem domains such that the uncertainty becomes manageable but still useful. Once these heuristics are recognized as a part of human reasoning it no longer appears illogical in the sense of being erratic, but rather more pragmatic and difficult to specify in terms of the logical inference mechanisms of traditional logic. Humans simplify decision making situations and use mental shortcuts to reach solutions which are satisfactory within constraints, but not necessarily optimal with respect to formal mathematical theory. Kahneman and Tversky (ibid) provide numerous examples in which subjects reach decisions which run counter to those reached by mathematical theories.

The research reported in this paper pursues the problem of developing representational and inference mechanisms for talking about human reasoning under conditions of uncertainty. The direction we have taken is based on the belief that methods which model the way people think under uncertainty may be used in the construction of flexible and more understandable computational reasoning systems. The model developed here involves collecting reasons for believing or disbelieving propositions as Cohen

57

(1983) does in his model of endorsement, and then qualifying these reasons by a measure of belief. In addition the belief measures can have varying degrees of certainty. The belief and certainty values can be used: (1) to determine how supportive a body of evidence for a particular hypothesis is and (2) to represent evidential relationships such as conflicts between decisions (Craddock 1986).

**2. BELIEF AND CERTAINTY**: We shall now proceed with defining a network containing propositions with beliefs and certainties, interconnected by arcs. First, let P be a set of cognitive units $P = \{n_1 ... n_m\}$. Each of these cognitive units may represent a proposition such as I LIKE MATH or relationships among objects such as one might employ in descriptions of visual scenes. Each $n_i$ has associated with it a belief strength $b_i$, which is a measure of the extent to which the cognitive unit is believable. The believability of $n_i$ is a measure of the completeness of the supporting evidence for $n_i$, not a measure of its incidence of occurrence or its possibility of occurrence. As $-1 \leq b_i \leq 1$, we may view the cognitive units as statements in a fuzzy propositional logic in which a belief of -1 indicates $n_i$ is false and a belief of +1 indicates $n_i$ is true.

In addition, each $n_i$ has associated with it a certainty, $c_i$ of the assignment of the value $b_i$, where $0 \leq c_i \leq 1$. The certainty of a belief value is defined as a measure of the reliability of the evidence which was used to calculate a particular belief (Hamburger, 1985). Thus each cognitive unit represents two distinct aspects of the *Rational* $R_i = (b_i, c_i)$ of $n_i$. To illustrate the distinction between the components, consider I LIKE MATH. This may have, for example $b_i = 0.4$ indicating moderately strong liking for mathematics [1]. On the other hand, the certainty $c_i$ of this value might be high or low, depending on the person's exposure to mathematics. Thus, it is important to note that the measure of a statement's certainty is not closely related to its belief. A statement can be highly believable but still be very uncertain. In a like fashion, a statement can be unbelievable but its incredibility may be very certain.

Any cognitive unit may endorse another. Each endorsement has associated with it a numeric value corresponding to the extent of the support between the units. If $n_i$ endorses $n_j$ then the support node $s_{ij}$ is the *support for the endorsement* where $-1 \leq s_{ij} \leq 1$. If $-1 \leq s_{ij} < 0$ then the endorsement $n_i$ for $n_j$ is said to be inhibitory and if $0 \leq s_{ij} \leq 1$ then it is said to be excitatory. The support nodes for the endorsement, $s_{ij}$ may be endorsed by other cognitive units. For example: I LIKE PSYCHOLOGY may endorse I LIKE COMPUTING but the support for the endorsement may be contingent on COMPUTATION MAY MODEL COGNITION(see figure 1). If COMPUTATION MAY MODEL COGNITION is false then the support for the endorsement will decrease (Craddock 1986).

Before going on to consider the mechanisms which permit dynamic evaluations and maintenance of belief based on these assigned values, it is useful to develop a network representation for the belief system. If we consider $P = \{n_1 ... n_m\}$ as the set of propositional nodes of the network, then we can define $S' = \{s_{ij} | n_i, n_j \in P, s_{ij} \neq 0\}$ as a subset of support nodes such that $n_i$ endorses $n_j$ with support $s_{ij}$, and $T = \{t_{s_{ij}} | s_{ij} \in S', n_k \in P\}$ as the other subset of support nodes such that $n_k$ endorses $s_{ij}$ with support $t_{s_{ij}}$. We can then define the network $N = <P, S>$ where $S = S' \cup T$ is a finit set of support nodes representing the arcs and P is a finit set of propositions.

**3. COMPUTING BELIEF AND CERTAINTY VALUES**: We wish to develop ways of computing the values of Rationale $R_i$ for a proposition $n_i$ on the basis of the endorsements available for that node. The *strength of an endorsement* between two nodes must be computed with consideration of $b_i$. For example, given the structure as shown in figure 2, if a person likes essays, ie. $b_i = .8$, then a net negative endorsement results for the node representing LIKING COMPUTING SCIENCE, but if the value of $b_i$ is low, a net positive endorsement would result. We can compute this endorsement strength as $e_{ij} = b_i s_{ij}$.

In this system, the belief strength of a node may be computed from the beliefs and certainties of its endorsements. We note, however, that it is not possible to compute this for all nodes in the network because to do so would require every node to have endorsements. Any node in the system, can be provided with an *Intuition* represented as $I_i = (b'_i, c'_i)$. This structure appears much the same as the

---

[1] This system does not distinguish between liking most mathematics, or liking all mathematics to some extent.

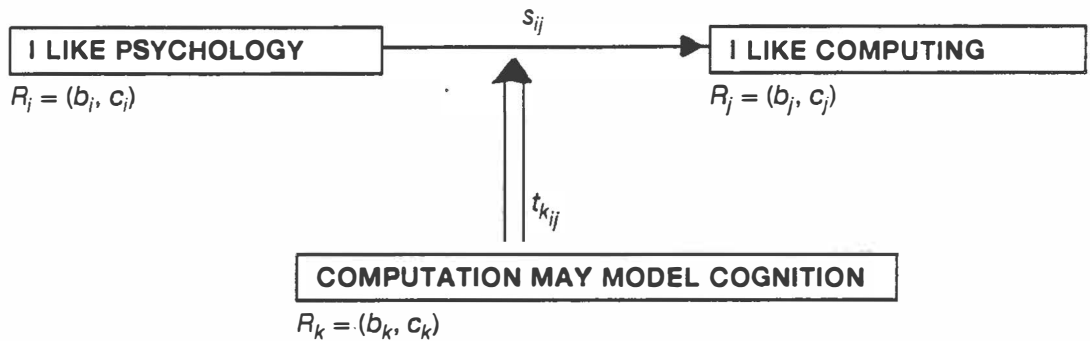

**Figure 1.** : An example of an endorsement which influences the strength of endorsement between two other nodes.

Rationale structure except that its values are never computed, but they remain available to take part in the computation of other beliefs. As we shall see later on, if a node has both a Rationale, and an Intuition, the system will check to verify that the values do not become inconsistent.

The computation of a belief value for a node is a function of the summary of the evidence and arguments. In such instances belief depends on: (1) measuring the varying contributions of the individual endorsements and (2) measuring the effects of interaction among the different endorsements. This interaction among a set of endorsements $\{n_1 \ldots n_m\}$ for $n_j$ depends on the relative importance of each endorsement defined as

$$r_{ij} = \frac{s_{ij}}{\sum_{k=1}^{m} |s_{kj}|} \qquad [1]$$

and the relative certainty defined as

$$rc_i = \frac{c_i}{c^*} \qquad [2]$$

where $c^*$ is max $\{c_i \ldots c_m\}$. In this manner we can define a measure of belief using a formula such as

$$b_j = \sum_{k=1}^{m} rc_k \times r_{kj} \times b_k \qquad [3]$$

Allowing the support of an endorsement to be endorsed enables us to deal with situations in which the endorsements $\{n_1 \ldots n_m\}$ for $n_j$ are mutually exclusive. $n_j$ will be given an endorsement of $b^* s_{kj}$ where $b^* = \max \{b_1 \ldots b_m\}$. The node $n_k$ will inhibit all the other endorsements by giving an inhibitory endorsement to their respective supports for the endorsement. This mutual exclusivity is somewhat similar to Rumelhart and Zipser's (1985) "Inhibitory Clusters" where nodes were arranged in groups in which the node with the highest activation inhibits the activations of all the other nodes.

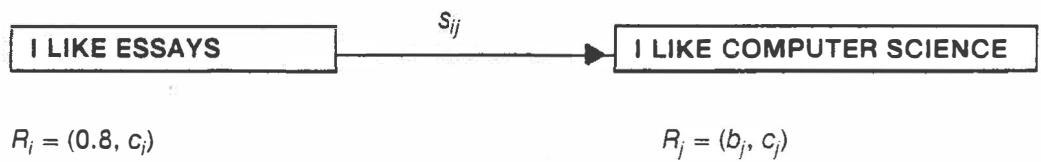

**Figure 2.** : An example of an endorsement which may have net positive or net negative support.

The certainty of a belief is calculated as a function of the agreement of the individual endorsement strengths with the final belief value calculated from them. Thus, belief must be calculated before certainty. The importance of the agreement is once again measured as a function of the relative support of the individual endorsements and their relative certainties. As these values increase so does the uncertainty associated with disagreement. Where $\{n_1 \ldots n_m\}$ are the endorsing nodes for $n_j$, this effect can be modelled in formulas such as:

$$c_j = 1 - \left[ \sum_{k \neq j, k=0}^{m} |(b_k - b_i)| \times r_{ki} \times rc_k \right] \quad [4]$$

**4. A NETWORK OF BELIEF STRUCTURES** The model described so far is similar to many existing connectionist models, particularly the spreading activation models of Anderson (1982), Rumelhart and McClelland (1982), and McClelland and Rumelhart (1983). However, because the model is intended to represent belief maintenance with uncertainty it differs in several important respects. First, the uncertainty of a proposition is represented numerically, as the values of $R_i$, and non-numerically, as the structure of endorsements. Second, once the endorsements have been collected they are subject to reasoning and natural heuristics to compute numeric values as depicted in formulae [1] to [4]. In contrast, most connectionist models ignore, or do not explicitly deal with the non-numeric representation of uncertainty, depending on numeric values alone which provide no evidence as to how they were calculated, what they actually represent, or how reliable they are.

Third, many recent spreading activation models (Anderson, 1982) propose that no limit be placed on the numeric "activation level" associated with a node. Instead, a decay function, dependent on the elapsed time since activation, is used to stop activation levels from increasing without bound. This allows the models to use the level of activation as an "associative relevancy" heuristic (Anderson, 1982) which states: The greater the activation of a node, the more relevant and closely associated it is to the problem. However, in this model we are representing belief, not activation, and belief can be bounded by truth, $b_i = 1$, and falsehood, $b_i = -1$. The relative importance of an endorsement, $r_{ij}$, and its relative certainty, $rc_i$, provide a measure similar to the "associative relevance" heuristic. Using a decay function can provide erroneous results in a belief maintenance system. Consider, for example, that doing WELL IN MATH endorses being a GOOD MATH STUDENT which in turn endorses taking a MATH MAJOR. In a model with a decay function the activation for being a MATH MAJOR will be less if activation has spread from the first endorsement than if it spread only from GOOD MATH STUDENT. This is because it takes longer for activation to spread through three links as opposed to two. This is counter-intuitive, because if GOOD MATH STUDENT is unsupported then it is not really as good a source of support. For this reason, this model uses inhibitory endorsements and bounded beliefs along with a relaxation algorithm to control the propagation and level of belief (Craddock 1986).

A further difference is that the proposed model can represent varying degrees of interaction between sources of evidence while models such as MYCIN (Shortcliffe, 1975) must make the assumption that all evidence is conditionally independent and that hypotheses are mutually exclusive. For example, the dynamic strengths of endorsement (see figure 1) allow us to represent evidence which is disjunctive, that is, strong belief may be propagated on the basis of only one of many supports. The endorsement with the greatest combination of relative certainty, belief and support inhibits the strengths of endorsement from the other nodes endorsing the decision so that they are effectively ignored. The amount of inhibition can be varied so that in instances where the endorsements are dependent no inhibition takes place. This is not to be confused with instances in which the proposition's belief is inhibitorly endorsed. To do so would indicate that its supporting evidence had been changed resulting in the unfortunate side-effect of altering the support the effected node would provide in an independent decision (Craddock 1986).

**5. HEURISTICS**: Within the model presented, uncertain knowledge is dealt with by using two distinct classes of heuristics. The first class of heuristics are used to determine which information is relevant to a decision. This class of heuristics is used to guide the collecting of endorsements. The second class of heuristics can then be used to evaluate the collected endorsements.

The first class of heuristics are influenced by the methods which humans appear to use while constraining and simplifying a problem; making the uncertain knowledge manageable but still useful. Of particular importance are Kahneman and Tversky's (1973, 1982a, b, c) heuristics of **relevance** and **availability**, and Anderson's (1982) previously mentioned heuristic of **associative relevance**. These heuristics make the common statement that *the only propositions which will be endorsed are those with endorsements*. It is safe to assume that those propositions which are not endorsed are either irrelevant or unavailable to the decision task. As mentioned in section 4 the available information with the greatest combined belief, relative certainty and relative importance is weighted with proportionate strength in the calculation of belief and certainty in formulae [3] and [4].

The second class of heuristics are influenced by the findings that bodies of evidence interact in complex ways. Among those modelled is Kahneman and Tversky's (ibid) heuristic of **adjustment and anchoring**, used to describe the human characteristic of making estimates from an initial value and then adjusting it to give the final answer. As mentioned in section 4 and formulae [3] and [4], endorsements with the greatest combined belief, relative certainty, and relative support, are weighted proportionately. The final belief values are thus biased or anchored towards these endorsements to model the phenomenon called adjustment.

In addition, the second class of heuristics attempt to model Kahneman and Tversky's (ibid) heuristic of **representativeness**. Kahneman and Tversky discovered that the similarity between data has a dramatic impact on how conclusions are weighted. As agreement between evidence increases so does the certainty of conclusions. As seen in formula [2], an endorsement with absolute uncertainty, $rc_i = 0$, will have no effect on the weighted sum of support in [3] and [4]. This heuristic is called the **ignorance** heuristic. In a similar fashion the **certainty** heuristic states that as the overall certainty of a set of endorsements decreases, i.e. as $\Sigma rc_i$ approaches 1, so does the resultant certainty, i.e. $1 - \Sigma |b_k - b_j| r_{kj} rc_k$ approaches 0. Finally, in formulae [4], certainty is inversely proportional to the disagreement between an endorsements belief, $b_i$, and the newly calculated belief, $b_j$, of the endorsed node. This allows us to model the **range** heuristic which states that certainty will be a maximum if all the endorsements agree in their new belief of the node, and the **resolution** heuristic, which states that as supporting evidence moves closer to the same value, disagreement will decrease and so will uncertainty. We shall now see what happens when endorsements are in complete disagreement.

**6. CONTRADICTIONS**: *Rational contradictions* among endorsements are defined as follows: If A is compelling evidence against $n_i$ but b is equally compelling evidence for $n_i$ then the endorsements for $n_i$ are inconsistent. In addition to a rational contraction an *intuitive contradiction* can also be defined: If the intuitive belief, $b'_i$ is not equal to the rational belief, $b_i$ then the two beliefs are inconsistent. Intuitive contradictions are useful for recognizing changes in belief through a knowledge base when knowledge is added and removed. In addition they can be used to control cycles which may force more global interpretations on input propositions. When cycles exist within a network $N = <P, S>$, belief and certainty values will only be calculated for nodes in a partial network $N' = P', S'$, where $P' \subseteq P$, and $S' \subseteq S \cap (P \times P')$, where there exists a node $n_i \in P - P'$ such that there is an elementary path between $n_i$ to $P'$ and $|I_i - R_i| > T_i$.

**CONCLUSIONS**: The model discussed in this paper seeks to develop representational and inference mechanisms capable of dealing with incomplete, inaccurate, and uncertain information. To this end, a model is proposed and heuristics are developed to collect and evaluate the endorsements for propositions in the network of beliefs. At the same time it is intended that the model represent at least some of the processes used in human reasoning. As the downfall of many expert systems with large data bases is that they can not determine which data is relevant to the problem, artificial intelligence may be able to use reasoning methods similar to humans to deal more effectively with uncertainty in decision making.

A second conclusion is that heuristics provide a very useful means of guiding the collection and evaluation of information. The constraints that these heuristics place on the decision making not only simplify the task, but allow uncertainty to remain useful. While the heuristics proposed in this paper are by no means exhaustive nor the formulae necessarily optimal, they do illustrate how heuristics might be incorporated into the decision making model in a straight forward fashion (Craddock 1986).

Finally, of major issue is the belief that numerical values, blindly tallied, are an inadequate representation of reasoning. Symbolic structures of support are necessary to specify how and why numeric beliefs are calculated. The incorporation of numerically qualified endorsements allows the model to not only provide numerical information, but it can also describe its reasoning process from start to finish in terms of reasons for and against a decision. The advantages of having a model whose reasoning can be readily understood are numerous if the model is to be used in situations where it is imperative that the justifications for a decision be made clear. In its present prototype form it is difficult to determine the extent to which the model may act as a model of human reasoning. It does, however, offer some attractive directions for further research.

## REFERENCES

Anderson, R, **The architecture of cognition.** Harvard University Press, 1982.

Cohen, R, "The use of heuristic knowledge in decision theory," Diss. Stanford University, 1983.

Craddock, A. J., "Modelling uncertainty in a knowledge base," MSc. Thesis, Queeen's University at Kingston, 1986.

Hamburger, H., **Combining uncertain estimates.** Uncertainty and Probability in Artificial Intelligence. August 14-16, 1985. UCLA: Los Angeles, California, 1985.

Kahneman D., and Tversky A., "Variants of uncertainty," **Cognition**, 11 (1982a) pp. 143 - 157.

Kahneman D., and Tversky A., **Judgment under uncertainty: Heuristics and Biases.** Cambridge University Press, 1982b.

Lee, W., **Decision theory and human behaviour.** New York: Wiley, 1969.

Lindley, D., **Making Decisions.** New York: Wiley-Interscience, 1971

McCarthy, J.. "Circumscription - A form of non-monotonic reasoning." **Artificial Intelligence**, 13 (1980), pp. 27 - 39.

McClelland, J. L., and Rumelhart, D. E., "An interactive activation model of context effects in letter perception: Part 1. An account of basic findings," **Psychological Review**, 88 (1981) pp. 375-407.

McDermott, D., and Doyle, J., "Non-monotonic Logic I," **Artificial Intelligence**, 13 (1980) pp. 41 - 72.

McDermott, D., "Non-monotonic modal theories," Report No. 174, Computer Science Dept., Yale University, 1980.

Nosteller, F., and Nogee, P., "An experimental measurement of utility," **Journal of Political Economy**, 59 (1951), pp. 371 - 404.

Reiter, R., "A logic for default reasoning," **Artificial Intelligence**, 13 (1980), pp. 81 - 132.

Rumelhart, D. E., and McClelland, J. L., "An interactive activation model of context effects in letter perception: Part 2. The contextual enhancement effect and some tests and extensions of the model," **Psychological Review**, 89 (1982), pp. 60-94.

Rumelhart, D. E. and Zipser D., "Feature Discovery by Competitive Learning," **Cognitive Science**, 9: 1985, pp. 75 - 112.

Schoemaker, P., **Experiments on decisions under risk: The expected utility hypothesis.** Boston: Martinus Nijhoff, 1980.

Shortcliffe. **Computer-based medical consultations: MYCIN.** New York: American Elsevier, 1975.

Zadeh L.A., "Commonsense knowledge representation based on Fuzzy Logic," **Computer**, Vol 16, 10 (1983), pp. 61-66.

Zadeh L.A., "Making computers think like people," **IEEE Spectrum**, Aug. 1984,